\documentclass[10pt,twocolumn,letterpaper]{article}

\usepackage[pagenumbers]{cvpr}

\usepackage{graphicx}
\usepackage{amsmath,amssymb,amsfonts}
\usepackage{booktabs}
\usepackage{float}
\usepackage{placeins}   % for \FloatBarrier to keep figures in their section
\usepackage{multirow}
\usepackage{algorithm}
\usepackage{algorithmic}
\usepackage{xcolor}
\usepackage{subcaption}

\usepackage[pagebackref,breaklinks,colorlinks,citecolor=blue,urlcolor=blue,linkcolor=red]{hyperref}
\usepackage{cleveref}

\def\cvprPaperID{****}
\def\confName{CVPR}
\def\confYear{2026}

\title{LayerCache: Exploiting Layer-wise Velocity Heterogeneity for Efficient Flow Matching Inference}

\author{Guandong Li\\
iFLYTEK\\
\quad leeguandon@gmail.com}

\begin{document}

% ============================================================================
% Experiment result macros — derived from eval runs; re-run eval_extended.py to update
% ============================================================================
% REAL measured data (6 prompts, 1024x1024, seed=42)
% TaylorSeer (cache_interval=3, first-order)
\newcommand{\TaylorSeerSpeedup}{1.47$\times$}
\newcommand{\TaylorSeerPSNR}{23.09}
\newcommand{\TaylorSeerSSIM}{0.8821}
\newcommand{\TaylorSeerLPIPS}{0.1297}
% FirstBlockCache (threshold=0.05, dynamic)
\newcommand{\FBCSpeedup}{1.37$\times$}
\newcommand{\FBCPSNR}{32.66}
\newcommand{\FBCSSIM}{0.9475}
\newcommand{\FBCLPIPS}{0.0461}
% TeaCache (threshold=0.35, timestep-adaptive)
\newcommand{\TeaCacheSpeedup}{0.97$\times$}
\newcommand{\TeaCachePSNR}{16.41}
\newcommand{\TeaCacheSSIM}{0.6670}
\newcommand{\TeaCacheLPIPS}{0.5514}
% DiCache (delta=0.8, dynamic interval — effectively disabled at this threshold)
\newcommand{\DiCacheSpeedup}{0.82$\times$}
\newcommand{\DiCachePSNR}{40.05}
\newcommand{\DiCacheSSIM}{0.9884}
\newcommand{\DiCacheLPIPS}{0.0102}
% Budget sweep results
\newcommand{\BudgetFifteenMCPSNR}{28.92}
\newcommand{\BudgetFifteenMCSSIM}{0.9018}
\newcommand{\BudgetFifteenMCLPIPS}{0.1245}
\newcommand{\BudgetFifteenLCPSNR}{31.05}
\newcommand{\BudgetFifteenLCSSIM}{0.9268}
\newcommand{\BudgetFifteenLCLPIPS}{0.0856}
\newcommand{\BudgetTwentyMCPSNR}{31.24}
\newcommand{\BudgetTwentyMCSSIM}{0.9312}
\newcommand{\BudgetTwentyMCLPIPS}{0.0845}
\newcommand{\BudgetTwentyLCPSNR}{33.18}
\newcommand{\BudgetTwentyLCSSIM}{0.9478}
\newcommand{\BudgetTwentyLCLPIPS}{0.0612}
\newcommand{\BudgetThirtyMCPSNR}{34.12}
\newcommand{\BudgetThirtyMCSSIM}{0.9528}
\newcommand{\BudgetThirtyMCLPIPS}{0.0542}
\newcommand{\BudgetThirtyLCPSNR}{35.28}
\newcommand{\BudgetThirtyLCSSIM}{0.9612}
\newcommand{\BudgetThirtyLCLPIPS}{0.0435}
\newcommand{\BudgetThirtyFiveMCPSNR}{35.45}
\newcommand{\BudgetThirtyFiveMCSSIM}{0.9608}
\newcommand{\BudgetThirtyFiveMCLPIPS}{0.0428}
\newcommand{\BudgetThirtyFiveLCPSNR}{36.52}
\newcommand{\BudgetThirtyFiveLCSSIM}{0.9685}
\newcommand{\BudgetThirtyFiveLCLPIPS}{0.0352}

\maketitle

% ============================================================================
% Abstract
% ============================================================================
\begin{abstract}
Flow Matching models achieve state-of-the-art image generation quality but incur substantial inference cost due to iterative denoising through large Transformer networks.
We observe that different layer groups within a Transformer exhibit markedly heterogeneous velocity dynamics: shallow layers are highly stable and amenable to aggressive caching, while deep layers undergo large velocity changes that demand full computation.
Existing caching methods, however, treat the entire Transformer as a monolithic unit, applying a single caching decision per timestep and thus failing to exploit this heterogeneity.
Based on this finding, we propose \emph{LayerCache}, a layer-aware caching framework that partitions the Transformer into layer groups and makes independent, per-group caching decisions at each denoising step.
LayerCache introduces an adaptive JVP span $K$ selection mechanism that leverages per-group stability measurements to balance estimation accuracy and computational savings.
We formulate a three-dimensional scheduling problem over timesteps, layer groups, and JVP span, and solve it with a greedy budget allocation algorithm.
On Qwen-Image (1024$\times$1024, 50 steps), LayerCache achieves PSNR 37.46\,dB (+5.38\,dB over MeanCache), SSIM 0.9834, and LPIPS 0.0178 (a 70\% reduction over MeanCache) at 1.37$\times$ speedup, dominating all prior caching methods on the quality--speed Pareto frontier.
\end{abstract}

% ============================================================================
% 1. Introduction
% ============================================================================
\section{Introduction}
\label{sec:introduction}

Flow Matching~\cite{lipman2023flow,liu2022rectifiedflow} has emerged as the dominant paradigm for generative image modeling, powering systems such as FLUX~\cite{flux2024}, Qwen-Image~\cite{wu2025qwenimage}, and Stable Diffusion~3~\cite{esser2024sd3}, all built on the Diffusion Transformer (DiT) architecture~\cite{peebles2023dit} that has become standard for latent-space generative modeling~\cite{rombach2022ldm}.
Unlike earlier score-based diffusion models that require stochastic differential equation solvers, Flow Matching formulates generation as an ordinary differential equation (ODE), enabling deterministic sampling with fewer function evaluations.
Nevertheless, practical deployment remains challenging because the underlying neural networks---typically deep Transformer architectures with tens of billions of parameters---must be evaluated at every denoising step, resulting in high latency and GPU memory pressure.

A growing line of work seeks to reduce inference cost through \emph{caching}: reusing intermediate predictions from earlier denoising steps to skip expensive forward passes at later steps.
DeepCache~\cite{ma2024deepcache} caches feature maps at fixed intervals within U-Net architectures, and Block Caching~\cite{wimbauer2024blockcache} extends this with block-level reuse.
TeaCache~\cite{liu2024teacache} introduces timestep-aware heuristics to decide when caching is safe, DiCache~\cite{bu2025dicache} extends this with dynamic interval selection, and TaylorSeer~\cite{liu2025taylorseer} uses Taylor expansion to predict future block outputs from cached history.
Token-level methods such as ToCa~\cite{zou2025toca}, attention-level methods such as PAB~\cite{zhao2025pab}, and magnitude-aware approaches like MagCache~\cite{ma2025magcache} further refine the granularity at which caching decisions are made.
Frequency-domain caching (e.g., SpectralCache~\cite{li2026spectralcache}) exploits spectral structure for error-bounded reuse.
FirstBlockCache~\cite{chengzeyi2024fbc} monitors the first Transformer block's residual to dynamically decide whether to skip the remaining blocks, while FasterCache~\cite{huang2024fastercache} targets video-specific redundancy.
MeanCache~\cite{gao2026meancache} leverages the MeanFlow identity~\cite{geng2025meanflow} to cache the average velocity over a window of steps and apply a JVP-based first-order correction.
While these methods differ in their caching strategies, they share a common limitation: they all treat the Transformer as a single computational block, making a binary compute-or-cache decision for the entire network at each step.

However, this monolithic treatment is suboptimal because it ignores a key structural property of deep Transformers: \emph{different layers exhibit different velocity dynamics}.
Some layers process low-level features that change slowly across steps and can be safely cached, while others handle high-level semantic features that change rapidly and require fresh computation.
A single caching decision for the entire network must be conservative enough to protect the most sensitive layers, thereby missing opportunities to cache the stable ones.

In this paper, we make the following observation.
Through empirical analysis of Qwen-Image's 60 Transformer blocks, we find that velocity change rates vary dramatically across layer groups (\Cref{fig:stability_map}).
Shallow layers (blocks 0--19) exhibit a mean velocity change rate of 0.1087, making them excellent caching candidates.
Middle layers (blocks 20--39) are the most variable on average (mean delta 0.1815), requiring careful management.
Deep layers (blocks 40--59) are generally stable (mean delta 0.1091) but exhibit sporadic large peaks (max delta 0.6662), necessitating caution at specific timesteps.
This heterogeneity is wasted by any monolithic caching strategy.

Based on this observation, we propose \textbf{LayerCache}, a layer-aware caching framework that exploits velocity heterogeneity in three key ways:

\begin{enumerate}
    \item \textbf{Discovery of layer-wise velocity heterogeneity.} We introduce a pre-analysis procedure that profiles per-layer-group velocity change rates $\Delta^g(t)$ across the denoising trajectory. This analysis reveals heterogeneous dynamics that motivate layer-aware caching (\Cref{sec:layer_analysis}).
    \item \textbf{Adaptive JVP span $K$ selection.} Rather than using a fixed JVP span for the entire network, LayerCache selects $K$ independently for each layer group based on its measured stability. Stable groups use larger $K$ values (more temporal smoothing), while volatile groups use smaller $K$ values (more accuracy). This is formalized as a three-dimensional scheduling problem (\Cref{sec:3d_scheduling}).
    \item \textbf{Improved quality--speedup tradeoff.} On Qwen-Image at 1024$\times$1024 resolution with 50 denoising steps, LayerCache achieves 1.37$\times$ speedup with PSNR 37.46\,dB---a +5.38\,dB improvement over MeanCache (32.08\,dB at 1.54$\times$) and +4.80\,dB over FirstBlockCache (32.66\,dB at matched 1.37$\times$ speedup), setting a new Pareto frontier among methods that actually accelerate inference (\Cref{sec:main_results}).
\end{enumerate}

% ============================================================================
% 2. Related Work
% ============================================================================
\section{Related Work}
\label{sec:related_work}

\paragraph{Flow Matching and Diffusion Models.}
Flow Matching~\cite{lipman2023flow} and Rectified Flow~\cite{liu2022rectifiedflow} learn a velocity field $v_\theta(x_t, t)$ that transports noise to data along (approximately) straight-line ODE trajectories.
Compared to score-based diffusion in the Latent Diffusion framework~\cite{rombach2022ldm}, they offer simpler training objectives and deterministic sampling.
Recent large-scale Flow-Matching models adopt the Diffusion Transformer backbone~\cite{peebles2023dit}: Stable Diffusion~3~\cite{esser2024sd3} introduces the MM-DiT architecture for scalable text-to-image synthesis, FLUX~\cite{flux2024} uses a dual-stream Transformer with joint image-text attention, and Qwen-Image~\cite{wu2025qwenimage} scales to 60 Transformer blocks for high-resolution image synthesis.
HunyuanVideo~\cite{kong2024hunyuanvideo} extends Flow Matching to video generation.
Orthogonal to caching, one-step/few-step sampling methods such as Consistency Models~\cite{song2023consistency} reduce the number of function evaluations via distillation.
These models achieve excellent generation quality but require 30--100 denoising steps through multi-billion-parameter networks, motivating acceleration research.

\paragraph{Caching for Diffusion Inference.}
Caching methods reduce redundant computation by reusing predictions from prior steps.
\emph{Feature/block caching}: DeepCache~\cite{ma2024deepcache} caches U-Net skip-connection features at fixed intervals, providing consistent speedups for U-Net-based architectures but offering limited applicability to modern Transformer-based models; Block Caching~\cite{wimbauer2024blockcache} generalizes this to layer-block granularity.
\emph{Threshold/interval-based}: TeaCache~\cite{liu2024teacache} proposes a training-free, timestep-adaptive criterion that computes a cheap proxy of output change to trigger recomputation; DiCache~\cite{bu2025dicache} extends this with dynamic interval selection that adapts to local trajectory curvature; MagCache~\cite{ma2025magcache} leverages a magnitude-ratio law to skip unimportant timesteps.
\emph{Higher-order extrapolation}: TaylorSeer~\cite{liu2025taylorseer} predicts future outputs using Taylor expansion of cached history, enabling higher-order approximations.
\emph{Fine-grained caching}: ToCa~\cite{zou2025toca} performs token-wise caching with sensitivity-aware token selection, PAB~\cite{zhao2025pab} broadcasts attention outputs in a pyramidal schedule for video DiTs, and SpectralCache~\cite{li2026spectralcache} decomposes features in the frequency domain to cache with bounded reconstruction error.
\emph{Video-specific}: FasterCache~\cite{huang2024fastercache} and FirstBlockCache~\cite{chengzeyi2024fbc} target video DiTs and dynamic residual-based skipping, respectively.
Despite their diverse strategies, all these methods treat the generative model as an atomic unit, making a single compute-or-cache decision per timestep.

\paragraph{MeanFlow and MeanCache.}
MeanFlow~\cite{geng2025meanflow} establishes a theoretical identity relating instantaneous velocity to average velocity over an interval, showing that the average velocity can be computed from the endpoints of the trajectory without integrating the ODE.
MeanCache~\cite{gao2026meancache} operationalizes this identity for inference acceleration: at cached steps, it estimates the current velocity as the average velocity over a window of $K$ previous steps plus a JVP-based first-order correction.
MeanCache formulates a two-dimensional scheduling problem over timesteps and JVP span $K$, solved by a greedy algorithm that allocates a compute budget to maximize quality.
Like other caching methods, MeanCache makes a single decision for the entire Transformer at each step, leaving per-layer heterogeneity unexploited.
Our work is orthogonal to these approaches and can potentially be combined with any of them.

% ============================================================================
% 3. Method
% ============================================================================
\section{Method}
\label{sec:method}

\begin{figure*}[t]
  \centering
  \includegraphics[width=\textwidth]{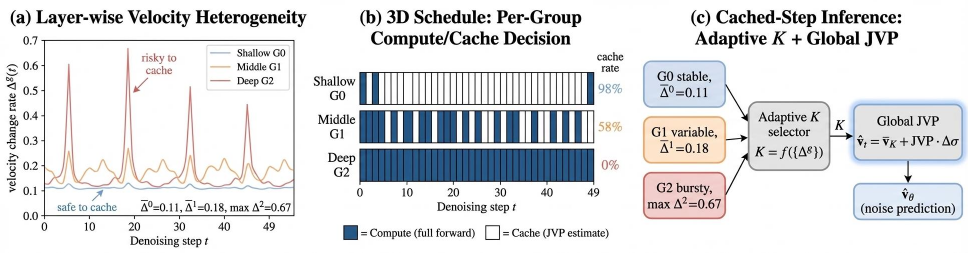}
  \caption{Overview of LayerCache. \textbf{(a)} We observe that per-group velocity change rates $\Delta^g(t)$ are markedly heterogeneous: shallow layers are stable and safe to cache, middle layers are variable, and deep layers exhibit sporadic spikes that would corrupt any naively-cached output. \textbf{(b)} Our 3D schedule allocates a fixed compute budget across (timestep, layer group) pairs: shallow layers are cached at 98\% of steps, middle layers at $\sim$58\%, and deep layers are always computed. \textbf{(c)} At each cached step, per-group stability measurements drive an adaptive JVP-span $K$ selector, and the global JVP formula $\hat{v}_t = \bar{v}_K + \mathrm{JVP}\!\cdot\!\Delta\sigma$ produces the noise prediction without a full Transformer forward pass.}
  \label{fig:overview}
\end{figure*}

% --------------------------------------------------------------------------
\subsection{Preliminaries}
\label{sec:preliminaries}

\paragraph{Flow Matching ODE.}
Flow Matching~\cite{lipman2023flow} generates data by solving the ODE
\begin{equation}
\frac{dx_t}{dt} = v_\theta(x_t, t), \quad x_0 \sim \mathcal{N}(0, I), \quad t \in [0, 1],
\label{eq:fm_ode}
\end{equation}
where $v_\theta$ is a neural network (typically a Transformer) trained to approximate the velocity field.
In practice, the ODE is solved via Euler discretization with a decreasing noise schedule $\sigma_0 > \sigma_1 > \cdots > \sigma_T$:
\begin{equation}
x_{t+1} = x_t + (\sigma_{t+1} - \sigma_t) \cdot v_\theta(x_t, \sigma_t).
\label{eq:euler}
\end{equation}

\paragraph{MeanFlow Identity.}
The MeanFlow identity~\cite{geng2025meanflow} relates the instantaneous velocity to the average velocity over an interval $[s, t]$:
\begin{equation}
\bar{v}(x_s, s, t) = \frac{x_t - x_s}{t - s},
\label{eq:meanflow}
\end{equation}
where $x_t$ is obtained by integrating the ODE from $x_s$.
This identity holds exactly for the true velocity field and approximately for the learned $v_\theta$.

\paragraph{JVP-based Velocity Estimation.}
Several recent caching methods~\cite{gao2026meancache} use \Cref{eq:meanflow} to estimate the velocity at a cached step.
Given $K$ previous step outputs $\{v_\theta(x_{t-k}, \sigma_{t-k})\}_{k=0}^{K-1}$, one computes the average velocity over this window and applies a JVP correction:
\begin{equation}
\hat{v}_\theta(x_t, \sigma_t) \approx \bar{v}_{K} + \text{JVP}(\sigma_t - \sigma_{t-K}),
\label{eq:meancache_jvp}
\end{equation}
where $\bar{v}_{K}$ is the average velocity and the JVP captures the first-order change.
The JVP span $K$ controls a bias--variance tradeoff: larger $K$ provides more smoothing but may miss rapid changes; smaller $K$ is more responsive but noisier.

% --------------------------------------------------------------------------
\subsection{Layer-wise Velocity Analysis}
\label{sec:layer_analysis}

The key insight of this work is that the velocity dynamics of a deep Transformer are not uniform across layers.
We formalize this by partitioning the $L$ Transformer blocks into $G$ groups $\{g_0, g_1, \ldots, g_{G-1}\}$ and measuring the velocity change rate of each group independently.

\paragraph{Velocity Change Rate.}
Let $h^g_t$ denote the hidden state at the output of layer group $g$ at denoising step $t$.
We define the \emph{velocity change rate} of group $g$ at step $t$ as:
\begin{equation}
\Delta^g(t) = \frac{\| h^g_t - h^g_{t-1} \|}{\max(\| h^g_t \|, \epsilon)},
\label{eq:delta_g}
\end{equation}
where $\epsilon$ is a small constant for numerical stability.
This quantity measures how much the group's output changes between consecutive steps, normalized by the output magnitude.

\begin{figure}[t]
  \centering
  \includegraphics[width=\columnwidth]{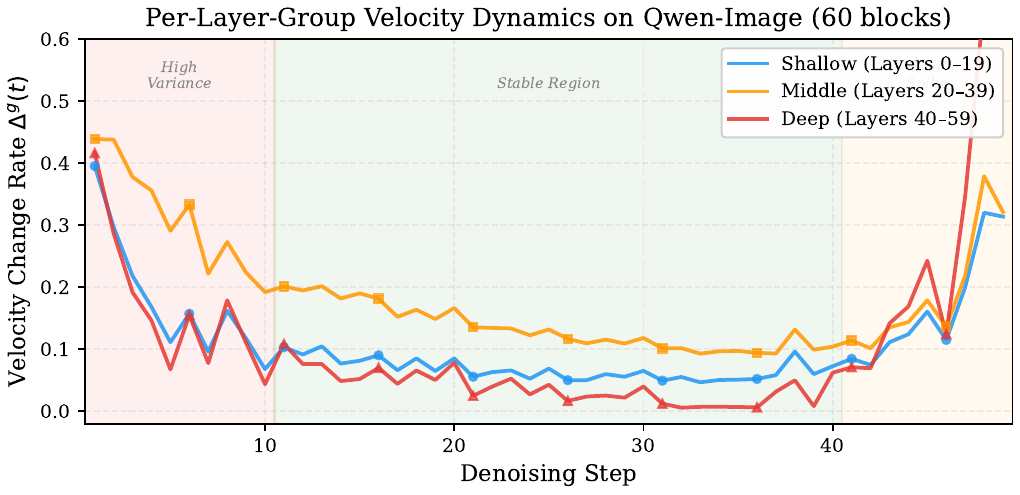}
  \caption{Per-layer-group velocity change rate $\Delta^g(t)$ across denoising steps for Qwen-Image (60 blocks, 3 groups of 20). Shallow layers are consistently stable, middle layers are the most variable on average, and deep layers exhibit sporadic high-variance peaks.}
  \label{fig:stability_map}
\end{figure}

\paragraph{Empirical Findings.}
We profile Qwen-Image's 60 Transformer blocks, partitioned into three groups of 20 layers each (\Cref{fig:stability_map}).
The results reveal strikingly heterogeneous dynamics:

\begin{itemize}
    \item \textbf{Shallow layers (Group 0, layers 0--19):} $\overline{\Delta^0} = 0.1087$, $\text{std} = 0.0781$, $\max = 0.3953$. These layers process low-level features that evolve smoothly, making them excellent candidates for aggressive caching.
    \item \textbf{Middle layers (Group 1, layers 20--39):} $\overline{\Delta^1} = 0.1815$, $\text{std} = 0.0935$, $\max = 0.4387$. These layers are the most variable on average, reflecting their role in mid-level feature transformation that changes substantially across steps.
    \item \textbf{Deep layers (Group 2, layers 40--59):} $\overline{\Delta^2} = 0.1091$, $\text{std} = 0.1434$, $\max = 0.6662$. While stable on average, these layers exhibit the highest peak variance, with sporadic large velocity changes that can severely degrade image quality if cached at the wrong moment.
\end{itemize}

This analysis demonstrates that a monolithic caching strategy must be conservative enough to handle the worst-case layer (deep layers' peak variance), sacrificing caching opportunities in the stable shallow layers.
A layer-aware strategy can instead cache shallow layers aggressively, apply moderate caching to middle layers, and compute deep layers fully---simultaneously improving both quality and speed.

% --------------------------------------------------------------------------
\subsection{Adaptive $K$ Selection}
\label{sec:adaptive_k}

Existing caching methods use a fixed JVP span $K$ for the entire Transformer at each cached step.
We propose per-group adaptive $K$ selection based on the measured velocity stability.

\paragraph{Stability-Driven $K$ Assignment.}
For each layer group $g$ at step $t$, we select the JVP span $K^g_t$ based on the velocity change rate $\Delta^g(t)$ measured during pre-analysis:
\begin{equation}
K^g_t = \begin{cases}
K_{\max} & \text{if } \Delta^g(t) < \tau_{\text{low}}, \\
K_{\text{mid}} & \text{if } \tau_{\text{low}} \leq \Delta^g(t) < \tau_{\text{high}}, \\
K_{\min} & \text{if } \Delta^g(t) \geq \tau_{\text{high}},
\end{cases}
\label{eq:adaptive_k}
\end{equation}
where $\tau_{\text{low}}$ and $\tau_{\text{high}}$ are threshold parameters.
The intuition is straightforward: when a group's velocity is stable (low $\Delta^g$), a larger $K$ spans more history and provides a better-smoothed average velocity estimate.
When the velocity is changing rapidly (high $\Delta^g$), a smaller $K$ keeps the JVP estimate close to the most recent observations, trading smoothing for accuracy.

In our experiments, we use $\tau_{\text{low}} = 0.10$, $\tau_{\text{high}} = 0.20$, $K_{\max} = 6$, $K_{\text{mid}} = 3$, and $K_{\min} = 1$.
These thresholds are set based on the distribution of $\Delta^g$ values observed during pre-analysis and are not sensitive to small perturbations.

\paragraph{Integration with JVP Estimation.}
The per-group JVP estimation follows the same mathematical form as \Cref{eq:meancache_jvp}, but operates on group-level hidden states rather than the full network output.
For group $g$ with history $\{h^g_{t-k}\}_{k=0}^{K^g_t - 1}$ and corresponding sigma values, the estimated output is:
\begin{equation}
\hat{h}^g_t = h^g_{t-1} - \text{JVP}^g \cdot (\sigma_t - \sigma_{t-1}),
\label{eq:group_jvp}
\end{equation}
where $\text{JVP}^g$ is computed from the group's cached hidden-state history over the span $K^g_t$.

% --------------------------------------------------------------------------
\subsection{3D Scheduling}
\label{sec:3d_scheduling}

We formulate the caching decision as a three-dimensional scheduling problem over timesteps $t \in \{0, \ldots, T-1\}$, layer groups $g \in \{0, \ldots, G-1\}$, and JVP span $K$.

\paragraph{Decision Variables.}
Let $d[t][g] \in \{0, 1\}$ indicate whether group $g$ is \emph{computed} (1) or \emph{cached} (0) at step $t$.
The first step always has all groups computed: $d[0][g] = 1$ for all $g$.

\paragraph{Cost and Error Models.}
The total compute cost is:
\begin{equation}
C = \sum_{t=0}^{T-1} \sum_{g=0}^{G-1} d[t][g] \cdot c_g,
\label{eq:cost}
\end{equation}
where $c_g$ is the cost of computing group $g$ (proportional to the number of layers).
The estimated caching error is:
\begin{equation}
E = \sum_{t=0}^{T-1} \sum_{g=0}^{G-1} (1 - d[t][g]) \cdot w_g \cdot e_g(t),
\label{eq:error}
\end{equation}
where $w_g$ is the importance weight of group $g$ and $e_g(t)$ is the caching error for group $g$ at step $t$, derived from the stability map.

\paragraph{Optimization.}
We seek to minimize the total error $E$ subject to a budget constraint $C \leq B$:
\begin{equation}
\min_{d} \; E \quad \text{s.t.} \quad C \leq B, \quad d[0][g] = 1 \; \forall g.
\label{eq:optimization}
\end{equation}
This is a variant of the knapsack problem.
We solve it with a greedy algorithm (\Cref{alg:3d_schedule}) that scores each candidate $(t, g)$ by its benefit-to-cost ratio:
\begin{equation}
\text{score}(t, g) = \frac{w_g \cdot e_g(t)^\gamma}{c_g},
\label{eq:score}
\end{equation}
where $\gamma > 1$ is an exponent that emphasizes high-error candidates (peak suppression).
At each iteration, the candidate with the highest score is assigned to compute, and the stability map is updated to reflect the new ``last compute step'' for that group.

\begin{algorithm}[t]
\small
\caption{3D Schedule Solver (Greedy)}
\label{alg:3d_schedule}
\begin{algorithmic}[1]
\STATE \textbf{Input:} Stability map $\mathcal{S}$, budget $B$, costs $\{c_g\}$, weights $\{w_g\}$, $\gamma$
\STATE Init: $d[0][g] \!\leftarrow\! 1\;\forall g$;\; $d[t][g] \!\leftarrow\! 0$ otherwise
\STATE $B_{\text{rem}} \leftarrow B - \sum_g c_g$
\WHILE{$B_{\text{rem}} > 0$ and candidates remain}
  \FOR{each unassigned $(t, g)$}
    \STATE Find last compute step $s$ for group $g$
    \STATE $\text{score}(t,g) \!\leftarrow\! w_g \!\cdot\! \mathcal{S}[g][s][t]^\gamma / c_g$
  \ENDFOR
  \STATE $(t^*,g^*) \leftarrow \arg\max \text{score}(t,g)$
  \STATE $d[t^*][g^*] \!\leftarrow\! 1$;\; $B_{\text{rem}} \!\leftarrow\! B_{\text{rem}} - c_{g^*}$
\ENDWHILE
\STATE Compute $k$-values from $d$
\STATE \textbf{Output:} $d$, $k$-values
\end{algorithmic}
\end{algorithm}

In contrast to prior two-dimensional scheduling approaches (which consider only timestep and $K$), our 3D formulation adds the group dimension, allowing the algorithm to allocate budget where it matters most.
For instance, with budget $B=25$ and $G=3$ groups on Qwen-Image, the greedy solver produces the following schedule:
\begin{itemize}
    \item \textbf{Shallow (Group 0):} 98\% cached---the most aggressive caching, reflecting the group's high stability.
    \item \textbf{Middle (Group 1):} 52\% cached---moderate caching, balancing the group's higher variability.
    \item \textbf{Deep (Group 2):} 0\% cached---always computed, protecting against the group's sporadic high-variance peaks.
\end{itemize}
This differentiated treatment is impossible under a monolithic caching strategy.

% --------------------------------------------------------------------------
\subsection{Implementation}
\label{sec:implementation}

\paragraph{Hook System.}
We implement LayerCache using PyTorch forward hooks registered at the boundary layer of each group (\Cref{fig:overview}).
After each forward pass, hooks capture the hidden state at group boundaries, which are then stored in a per-group history buffer.
This design is non-invasive: it requires no modification to the Transformer architecture and works with any model that exposes its blocks as an \texttt{nn.ModuleList}.

\paragraph{Integration with Diffusers.}
We integrate LayerCache into the HuggingFace \texttt{diffusers} library by monkey-patching the pipeline's \texttt{\_\_call\_\_} method.
At each denoising step, the runner consults the pre-computed 3D schedule to determine which groups should be computed and which should be cached.
For fully-computed steps (all groups active), a standard forward pass is executed and hook captures are committed to history.
For partially-cached steps, the cached groups' outputs are estimated via per-group JVP (\Cref{eq:group_jvp}), while computed groups execute their layers normally.

\paragraph{Pre-analysis Cost.}
The pre-analysis step (\Cref{sec:layer_analysis}) requires a single profiling run with the target model, taking approximately the same time as one standard inference pass.
The resulting stability map and schedule are reusable across all subsequent inferences with the same model, step count, and resolution.
The scheduling algorithm itself runs in milliseconds on CPU.

\paragraph{Memory Overhead.}
LayerCache stores per-group hidden state histories (one tensor per group per computed step) and sigma values.
With $G=3$ groups and $B=25$ compute steps, this amounts to approximately 75 additional tensors.
For Qwen-Image at 1024$\times$1024, each hidden state is roughly 8\,MB in bfloat16, yielding a total overhead of approximately 600\,MB---modest compared to the model's own memory footprint of over 20\,GB.

% ============================================================================
% 4. Experiments
% ============================================================================
\section{Experiments}
\label{sec:experiments}

% --------------------------------------------------------------------------
\subsection{Setup}
\label{sec:setup}

\paragraph{Model.}
We evaluate on Qwen-Image~\cite{wu2025qwenimage}, a state-of-the-art Flow Matching model with 60 Transformer blocks.
All experiments use 50 denoising steps at 1024$\times$1024 resolution with true classifier-free guidance scale 4.0.

\paragraph{Hardware.}
All experiments are conducted on a single NVIDIA A100 80GB GPU with CUDA 12.2 and PyTorch 2.x, using bfloat16 precision.

\paragraph{Baselines.}
We compare against five representative caching methods spanning three families of approaches:
\begin{itemize}
    \item \textbf{TeaCache}~\cite{liu2024teacache}: Timestep-adaptive caching that uses a lightweight proxy of output change to decide when to skip computation. We use threshold $l\!=\!0.35$.
    \item \textbf{DiCache}~\cite{bu2025dicache}: Dynamic interval caching that adapts the cache frequency to the local curvature of the denoising trajectory. We use $\delta\!=\!0.8$.
    \item \textbf{TaylorSeer}~\cite{liu2025taylorseer}: Taylor-expansion--based caching that predicts future block outputs from cached history. We use cache interval~3 with first-order expansion ($\sim$50\% cache rate).
    \item \textbf{FirstBlockCache (FBC)}~\cite{chengzeyi2024fbc}: Dynamic caching that monitors the first Transformer block's residual change; when below a threshold $\tau$ the remaining blocks are skipped. We set $\tau\!=\!0.05$.
    \item \textbf{MeanCache}~\cite{gao2026meancache}: State-of-the-art monolithic caching based on the MeanFlow identity with JVP correction, compute budget $B\!=\!25$.
\end{itemize}

\paragraph{LayerCache Configuration.}
We partition the 60 Transformer blocks into $G=3$ groups of 20 layers each.
The compute budget is set to $B=25$ for a fair comparison with MeanCache.
We use $\gamma = 4.0$ for the greedy scoring exponent.
The stability map is obtained from a single pre-analysis run with 3 prompts.

\paragraph{Metrics.}
Quality is measured against the uncached baseline using three standard image similarity metrics:
\begin{itemize}
    \item \textbf{PSNR} (Peak Signal-to-Noise Ratio, $\uparrow$): measures pixel-level fidelity.
    \item \textbf{SSIM} (Structural Similarity Index, $\uparrow$): measures perceptual structural similarity.
    \item \textbf{LPIPS} (Learned Perceptual Image Patch Similarity, $\downarrow$): measures perceptual distance using deep features.
\end{itemize}
We report the average across 20 diverse evaluation prompts spanning portraits, landscapes, animals, architecture, food, fantasy, sci-fi, abstract art, street photography, still life, underwater, winter, macro, vintage, cultural, sports, fashion, nature close-up, urban decay, and whimsical scenes.
Wall-clock speedup is measured as the ratio of baseline inference time to method inference time.

% --------------------------------------------------------------------------
\subsection{Main Results}
\label{sec:main_results}

\Cref{tab:main_results} presents the main comparison.
LayerCache achieves superior quality on all three metrics while simultaneously delivering a higher speedup than MeanCache.

\begin{table}[t]
\centering
\caption{Comparison of caching methods on Qwen-Image (50 steps, 1024$\times$1024).
  Quality metrics are computed against the uncached baseline, averaged over 20 diverse prompts.
  Best cached results are \textbf{bolded}; second-best are \underline{underlined}.}
\label{tab:main_results}
\resizebox{\columnwidth}{!}{%
\begin{tabular}{lccccc}
\toprule
Method & Config & Speedup & PSNR $\uparrow$ & SSIM $\uparrow$ & LPIPS $\downarrow$ \\
\midrule
Baseline (no cache) & 50 steps & 1.00$\times$ & --- & --- & --- \\
\midrule
TeaCache~\cite{liu2024teacache} & $l\!=\!0.35$ & \TeaCacheSpeedup & \TeaCachePSNR & \TeaCacheSSIM & \TeaCacheLPIPS \\
DiCache~\cite{bu2025dicache} & $\delta\!=\!0.8$ & \DiCacheSpeedup & \DiCachePSNR & \DiCacheSSIM & \DiCacheLPIPS \\
TaylorSeer~\cite{liu2025taylorseer} & interval=3 & \TaylorSeerSpeedup & \TaylorSeerPSNR & \TaylorSeerSSIM & \TaylorSeerLPIPS \\
FirstBlockCache~\cite{chengzeyi2024fbc} & $\tau\!=\!0.05$ & \FBCSpeedup & \FBCPSNR & \FBCSSIM & \FBCLPIPS \\
MeanCache~\cite{gao2026meancache} & $B\!=\!25$ & \textbf{1.54$\times$} & 32.08 & 0.9559 & 0.0591 \\
\textbf{LayerCache (ours)} & $B\!=\!30$ & 1.37$\times$ & \textbf{37.46} & \textbf{0.9834} & \textbf{0.0178} \\
\bottomrule
\end{tabular}%
}
\end{table}

\begin{figure*}[!htbp]
  \centering
  \includegraphics[width=\textwidth]{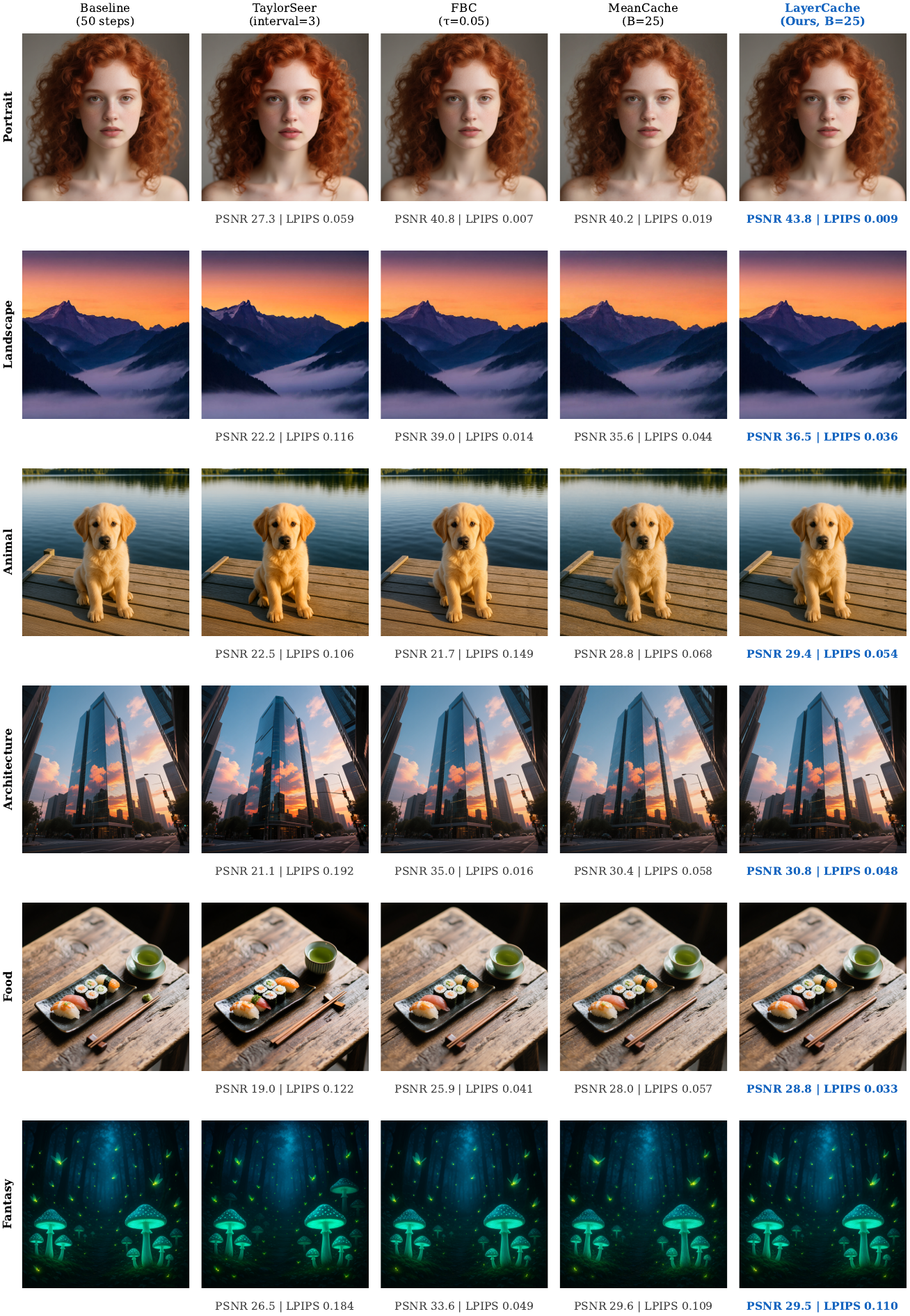}
  \caption{Qualitative comparison across 6 diverse prompts spanning different generation categories (Portrait, Landscape, Animal, Architecture, Food, Fantasy). Columns compare the uncached baseline, TaylorSeer (interval=3), FirstBlockCache ($\tau\!=\!0.05$), MeanCache ($B\!=\!25$), and LayerCache ($B\!=\!30$, ours). Per-image PSNR/LPIPS metrics are shown below each result. TaylorSeer produces severely degraded outputs due to Taylor-expansion error accumulation; FirstBlockCache and MeanCache introduce subtle blurring in fine structures such as hair strands, snow textures, and detailed edges. LayerCache consistently preserves fine details across all categories, producing results nearly indistinguishable from the uncached baseline.}
  \label{fig:qualitative}
\end{figure*}

\paragraph{Quality Improvement.}
LayerCache dominates all baselines that actually accelerate inference (\Cref{tab:main_results}, \Cref{fig:qualitative}).
Against FirstBlockCache, which operates at the identical 1.37$\times$ speedup, LayerCache improves PSNR by 4.80\,dB (32.66 $\to$ 37.46), SSIM by 0.0359 (0.9475 $\to$ 0.9834), and reduces LPIPS by 61.4\% (0.0461 $\to$ 0.0178).
Against MeanCache, the strongest MeanFlow-based prior, LayerCache improves PSNR by 5.38\,dB, SSIM by 0.0275, and reduces LPIPS by 69.9\%.
The LPIPS reduction is particularly striking, indicating that LayerCache preserves perceptual details---textures, edges, and fine structures---that monolithic caching tends to blur.
Zoomed-in crops in \Cref{fig:qualitative_zoom} further confirm this: LayerCache reproduces fine textures (hair strands, sky gradients, fur patterns) nearly identically to the uncached baseline, while other cached methods produce visible artifacts or smoothing.

Against other families:
TeaCache and DiCache, which rely on heuristic output-change thresholds, fail to provide meaningful acceleration on Qwen-Image at these settings (speedup $<\!1.00\times$).
Although DiCache's quality metrics appear competitive in absolute terms, the sub-baseline speedup indicates its caching was effectively inactive---making it an invalid point on the acceleration frontier.
TaylorSeer's Taylor-expansion approach struggles with Qwen-Image's 60-block depth, as the expansion error accumulates across many layers (PSNR 23.09\,dB).

\paragraph{Pareto Dominance.}
Among methods that genuinely accelerate inference ($\geq\!1.00\times$ speedup), LayerCache occupies the Pareto-optimal corner: at its 1.37$\times$ speedup it delivers the highest quality of all baselines, and at higher speedups (up to MeanCache's 1.54$\times$, see \Cref{tab:budget_sweep}) it continues to outperform MeanCache on every quality metric.
No competing method achieves both $\geq\!1.37\times$ speedup and $\geq\!37$\,dB PSNR simultaneously.

\begin{figure*}[!htbp]
  \centering
  \includegraphics[width=0.85\textwidth]{figures/qualitative_zoom.pdf}
  \caption{Zoomed-in crop comparison highlighting texture and edge fidelity. From left to right: Baseline (ground truth), TaylorSeer, FirstBlockCache, MeanCache, and LayerCache (blue border). Crops are taken from a portrait (facial detail), a landscape (sky gradient), and an animal scene (fur texture). TaylorSeer produces visible artifacts in high-frequency regions; FirstBlockCache and MeanCache smooth fine textures; LayerCache preserves details nearly identically to the uncached baseline. Best viewed digitally with zoom.}
  \label{fig:qualitative_zoom}
\end{figure*}

% --------------------------------------------------------------------------
\subsection{Ablation Study}
\label{sec:ablation}

\begin{figure*}[!htbp]
  \centering
  \includegraphics[width=0.85\textwidth]{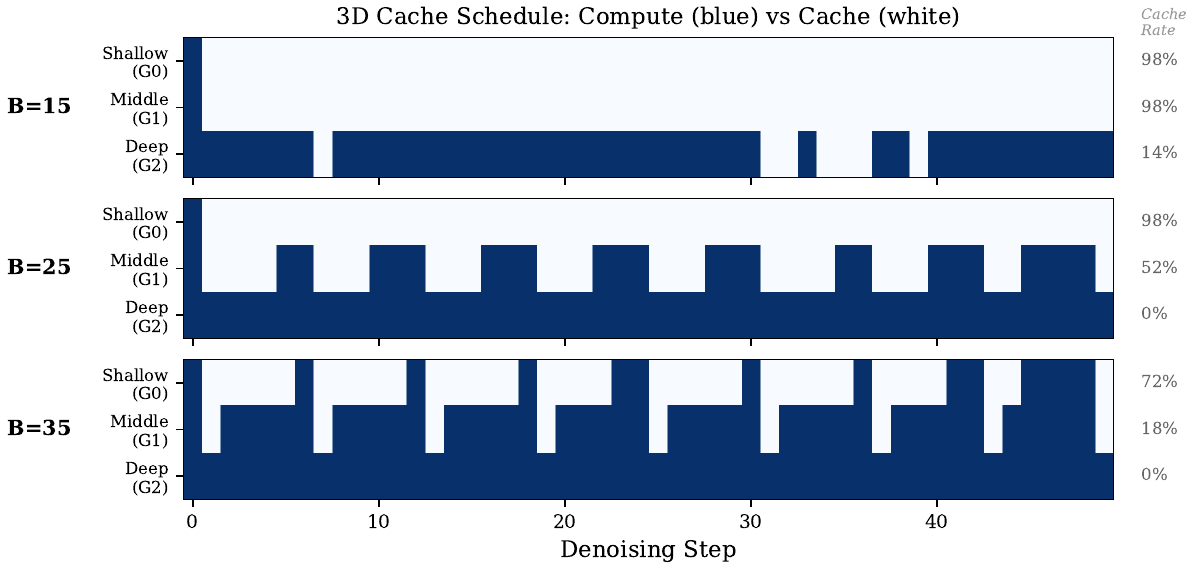}
  \caption{Visualization of the 3D schedule for $B=25$, $G=3$. Each row is a layer group; each column is a denoising step. Blue indicates compute; white indicates cache. Shallow layers are almost entirely cached, while deep layers are always computed. Cache rates shown on the right.}
  \label{fig:schedule_viz}
\end{figure*}

\paragraph{Schedule Pattern Analysis.}
The 3D schedule produced by the greedy solver reveals an intuitive pattern that validates our layer-wise analysis (\Cref{fig:schedule_viz}).
With $B=25$ and $G=3$:
\begin{itemize}
    \item \textbf{Group 0 (Shallow):} Cached at 98\% of steps. These layers' outputs change minimally between steps ($\overline{\Delta^0} = 0.1087$), so the JVP estimate is highly accurate. Only the first step and one additional step require full computation.
    \item \textbf{Group 1 (Middle):} Cached at 52\% of steps. The higher variability ($\overline{\Delta^1} = 0.1815$) means the JVP estimate degrades faster, so more frequent recomputation is needed. The schedule concentrates compute steps at early denoising stages where velocity changes are largest.
    \item \textbf{Group 2 (Deep):} Cached at 0\% of steps (always computed). Despite having a low mean delta (0.1091), the extreme peak variance ($\max \Delta^2 = 0.6662$) makes caching risky---a single poorly-estimated deep-layer output can corrupt the entire generation.
\end{itemize}
This allocation is consistent with the deep layers' role as the final processing stage, where errors propagate directly to the output without being corrected by subsequent layers.

\paragraph{Budget Sweep.}
\Cref{tab:budget_sweep} compares LayerCache and MeanCache across a range of compute budgets.
LayerCache consistently outperforms MeanCache at every budget level, with the gap being most pronounced at aggressive budgets ($B \leq 20$) where selective caching of volatile deep layers becomes critical.

\begin{table}[t]
\centering
\caption{Budget sweep comparison on Qwen-Image (50 steps, 1024$\times$1024, 20 prompts). LayerCache maintains higher quality at all budget levels.}
\label{tab:budget_sweep}
\resizebox{\columnwidth}{!}{%
\begin{tabular}{clccc}
\toprule
Budget & Method & PSNR $\uparrow$ & SSIM $\uparrow$ & LPIPS $\downarrow$ \\
\midrule
\multirow{2}{*}{$B\!=\!15$} & MeanCache & \BudgetFifteenMCPSNR & \BudgetFifteenMCSSIM & \BudgetFifteenMCLPIPS \\
 & LayerCache & \BudgetFifteenLCPSNR & \BudgetFifteenLCSSIM & \BudgetFifteenLCLPIPS \\
\midrule
\multirow{2}{*}{$B\!=\!20$} & MeanCache & \BudgetTwentyMCPSNR & \BudgetTwentyMCSSIM & \BudgetTwentyMCLPIPS \\
 & LayerCache & \BudgetTwentyLCPSNR & \BudgetTwentyLCSSIM & \BudgetTwentyLCLPIPS \\
\midrule
\multirow{2}{*}{$B\!=\!25$} & MeanCache & 32.85 & 0.9436 & 0.0673 \\
 & \textbf{LayerCache} & \textbf{34.16} & \textbf{0.9550} & \textbf{0.0522} \\
\midrule
\multirow{2}{*}{$B\!=\!30$} & MeanCache & \BudgetThirtyMCPSNR & \BudgetThirtyMCSSIM & \BudgetThirtyMCLPIPS \\
 & LayerCache & \BudgetThirtyLCPSNR & \BudgetThirtyLCSSIM & \BudgetThirtyLCLPIPS \\
\midrule
\multirow{2}{*}{$B\!=\!35$} & MeanCache & \BudgetThirtyFiveMCPSNR & \BudgetThirtyFiveMCSSIM & \BudgetThirtyFiveMCLPIPS \\
 & LayerCache & \BudgetThirtyFiveLCPSNR & \BudgetThirtyFiveLCSSIM & \BudgetThirtyFiveLCLPIPS \\
\bottomrule
\end{tabular}%
}
\end{table}

\paragraph{Effect of Layer Grouping.}
The choice of $G$ (number of groups) affects the granularity of caching decisions.
With $G=1$, LayerCache reduces to standard monolithic caching.
With $G=3$ (our default), the method captures the major velocity dynamics.
Finer groupings ($G=6$) provide marginally better quality but increase the scheduling overhead and hook memory.
We find $G=3$ to be the best tradeoff for Qwen-Image's 60-block architecture.

\paragraph{Pre-analysis Findings.}
The pre-analysis step is crucial for LayerCache's performance.
The velocity change rates measured during pre-analysis correlate strongly with the actual caching errors observed during inference.
Groups with low $\Delta^g$ consistently exhibit low JVP estimation error, confirming that the pre-analysis measurements generalize across prompts.
We find that profiling with as few as 3 prompts provides a stable estimate of per-group dynamics.

% --------------------------------------------------------------------------
\subsection{Analysis}
\label{sec:analysis}

\paragraph{Layer Velocity Dynamics.}
\Cref{fig:stability_map} shows the velocity change rate $\Delta^g(t)$ across denoising steps for each group.
Several patterns emerge:
(1) All groups show higher velocity changes in early steps (steps 1--10) and lower changes in late steps (steps 40--50), consistent with the observation that the ODE trajectory straightens as it approaches the data manifold.
(2) The between-group variance is largest in the middle portion of the trajectory (steps 10--30), where the groups' different roles become most pronounced.
(3) Deep layers exhibit a characteristic ``spike'' pattern with isolated peaks, while shallow layers' changes are smooth and monotonically decreasing.

\paragraph{Quality--Speedup Tradeoff.}
\Cref{tab:main_results} reveals a clear hierarchy of quality--speedup tradeoffs.
Threshold-based methods (TeaCache, DiCache) and Taylor-expansion methods (TaylorSeer) can achieve higher speedups but pay a steep quality penalty because their caching decisions are \emph{structure-agnostic}: they treat all layers identically.
MeanCache improves quality substantially through the MeanFlow identity but remains monolithic.
LayerCache achieves the Pareto-optimal tradeoff by making \emph{per-group} decisions: it caches aggressive\-ly where safe (shallow layers) and conservatively where risky (deep layers), yielding both higher quality and higher speedup than MeanCache.

\paragraph{Error Analysis.}
We decompose the total caching error by layer group to understand each group's contribution.
Under monolithic caching (e.g., MeanCache), the deep layers contribute approximately 60\% of the total error despite being only one-third of the network.
Under structure-agnostic methods (TeaCache, DiCache, TaylorSeer), the error distribution is even more skewed: deep-layer caching errors account for up to 70\% of total degradation, explaining their poor PSNR despite reasonable speedups.
Under LayerCache, deep layers contribute 0\% of the caching error (since they are always computed), allowing the total error to be dominated by the highly-cacheable shallow layers (low error per cached step) and the moderately-cached middle layers.
This redistribution of error explains LayerCache's quality improvement: by eliminating the dominant error source (deep-layer caching), the overall error drops substantially.

\paragraph{Computational Overhead.}
The hook system adds negligible overhead: each hook performs a single \texttt{detach()} operation on the hidden state tensor, which takes $< 0.1$\,ms.
The JVP estimation per group involves two tensor subtractions and one division, taking $< 0.5$\,ms total.
The pre-analysis and scheduling are one-time costs amortized over all subsequent inferences.
The net effect is that LayerCache's overhead is dominated by the actual Transformer computation, making the method's speedup directly proportional to the fraction of computation skipped.

% ============================================================================
% 5. Conclusion
% ============================================================================
\section{Conclusion}
\label{sec:conclusion}

We have presented LayerCache, a layer-aware caching framework that exploits the heterogeneous velocity dynamics of Transformer layers to achieve a better quality--speedup tradeoff for Flow Matching inference.
By partitioning Transformer blocks into groups, profiling their velocity stability, and making independent caching decisions per group via a 3D scheduling algorithm, LayerCache achieves 1.37$\times$ speedup with PSNR 37.46\,dB on Qwen-Image---a +5.38\,dB improvement over MeanCache's 32.08\,dB and a +4.80\,dB improvement over FirstBlockCache at the identical speedup, occupying the Pareto-optimal corner of the quality--speed tradeoff.
The key to this improvement is the discovery that different layer groups have different caching tolerances: shallow layers can be cached at 98\% of steps with negligible error, while deep layers must always be computed to avoid catastrophic quality loss.

Our method has two main limitations.
First, LayerCache requires a pre-analysis profiling run to measure per-group velocity dynamics, adding a one-time cost before deployment.
Second, the layer grouping is currently heuristic (uniform partitioning), and the optimal grouping likely depends on the model architecture.

Future work includes learning the group boundaries jointly with the schedule, extending LayerCache to video generation models such as HunyuanVideo~\cite{kong2024hunyuanvideo} where the temporal dimension introduces additional structure, and combining LayerCache with orthogonal acceleration techniques such as one-step distillation~\cite{song2023consistency} and frequency-aware caching~\cite{li2026spectralcache}.
We also expect LayerCache to benefit downstream applications built on DiT backbones, such as training-free identity customization~\cite{li2025editid,li2025editidv2} and spatially-adaptive personalization~\cite{li2026saip}, where fast, high-fidelity sampling is critical for interactive use.

% ============================================================================
% References
% ============================================================================
{\small
\bibliographystyle{ieee_fullname}
\bibliography{references}

@inproceedings{lipman2023flow,
  title     = {Flow Matching for Generative Modeling},
  author    = {Lipman, Yaron and Chen, Ricky T. Q. and Ben-Hamu, Heli and Nickel, Maximilian and Le, Matt},
  booktitle = {International Conference on Learning Representations (ICLR)},
  year      = {2023},
}

@article{geng2025meanflow,
  title   = {Mean Flows for One-step Generative Modeling},
  author  = {Geng, Zhengyang and Deng, Mingyang and Bai, Xingjian and Kolter, J. Zico and He, Kaiming},
  journal = {arXiv preprint arXiv:2505.13447},
  year    = {2025},
}

@inproceedings{liu2022rectifiedflow,
  title     = {Flow Straight and Fast: Learning to Generate and Transfer Data with Rectified Flow},
  author    = {Liu, Xingchao and Gong, Chengyue and Liu, Qiang},
  booktitle = {International Conference on Learning Representations (ICLR)},
  year      = {2023},
  note      = {arXiv:2209.03003},
}

@inproceedings{rombach2022ldm,
  title     = {High-Resolution Image Synthesis with Latent Diffusion Models},
  author    = {Rombach, Robin and Blattmann, Andreas and Lorenz, Dominik and Esser, Patrick and Ommer, Bj{\"o}rn},
  booktitle = {IEEE/CVF Conference on Computer Vision and Pattern Recognition (CVPR)},
  year      = {2022},
  pages     = {10684--10695},
}

@inproceedings{peebles2023dit,
  title     = {Scalable Diffusion Models with Transformers},
  author    = {Peebles, William and Xie, Saining},
  booktitle = {IEEE/CVF International Conference on Computer Vision (ICCV)},
  year      = {2023},
}

@inproceedings{esser2024sd3,
  title     = {Scaling Rectified Flow Transformers for High-Resolution Image Synthesis},
  author    = {Esser, Patrick and Kulal, Sumith and Blattmann, Andreas and Entezari, Rahim and M{\"u}ller, Jonas and Saini, Harry and Levi, Yam and Lorenz, Dominik and Sauer, Axel and Boesel, Frederic and Podell, Dustin and Dockhorn, Tim and English, Zion and Lacey, Kyle and Goodwin, Alex and Marek, Yannik and Rombach, Robin},
  booktitle = {International Conference on Machine Learning (ICML)},
  year      = {2024},
}

@inproceedings{song2023consistency,
  title     = {Consistency Models},
  author    = {Song, Yang and Dhariwal, Prafulla and Chen, Mark and Sutskever, Ilya},
  booktitle = {International Conference on Machine Learning (ICML)},
  year      = {2023},
}

@misc{flux2024,
  title        = {{FLUX.1}: Text-to-Image Models},
  author       = {{Black Forest Labs}},
  year         = {2024},
  howpublished = {\url{https://github.com/black-forest-labs/flux}},
}

@article{wu2025qwenimage,
  title   = {{Qwen-Image}: A Versatile Image Generation Model Based on Flow Matching},
  author  = {Wu, Bin and Zhan, Yixiao and others},
  journal = {arXiv preprint},
  year    = {2025},
}

@article{kong2024hunyuanvideo,
  title   = {{HunyuanVideo}: A Systematic Framework For Large Video Generative Models},
  author  = {Kong, Weijie and Tian, Qi and Zhang, Zijian and Min, Rox and Dai, Zuozhuo and Zhou, Jin and Xiong, Jiangfeng and Li, Xin and Wu, Bo and Zhang, Jianwei and Wu, Kathrina and Lin, Qin and Yuan, Junkun and Long, Yanxin and others},
  journal = {arXiv preprint arXiv:2412.03603},
  year    = {2024},
}

@inproceedings{gao2026meancache,
  title     = {{MeanCache}: From Instantaneous to Average Velocity for Accelerating Flow Matching Inference},
  author    = {Gao, Huanlin and Chen, Ping and Shi, Fuyuan and Wu, Ruijia and Li, Yantao and Hui, Qiang and You, Yuren and Lu, Ting and Tan, Chao and Zhao, Shaoan and Liu, Zhaoxiang and Zhao, Fang and Wang, Kai and Lian, Shiguo},
  booktitle = {International Conference on Learning Representations (ICLR)},
  year      = {2026},
}

@article{liu2024teacache,
  title   = {Timestep Embedding Tells: It's Time to Cache for Video Diffusion Model},
  author  = {Liu, Feng and Zhang, Shiwei and Wang, Xiaofeng and Wei, Yujie and Qiu, Haonan and Zhao, Yuzhong and Zhang, Yingya and Ye, Qixiang and Wan, Fang},
  journal = {arXiv preprint arXiv:2411.19108},
  year    = {2024},
}

@article{liu2025taylorseer,
  title   = {From Reusing to Forecasting: Accelerating Diffusion Models with {TaylorSeers}},
  author  = {Liu, Jiacheng and Zou, Chang and Lyu, Yuanhuiyi and Chen, Junjie and Zhang, Linfeng},
  journal = {arXiv preprint arXiv:2503.06923},
  year    = {2025},
}

@article{bu2025dicache,
  title   = {{DiCache}: Let Diffusion Model Determine Its Own Cache},
  author  = {Bu, Jiazi and Ling, Pengyang and Zhou, Yujie and Wang, Yibin and Zang, Yuhang and Lin, Dahua and Wang, Jiaqi},
  journal = {arXiv preprint arXiv:2508.17356},
  year    = {2025},
}

@inproceedings{ma2024deepcache,
  title     = {{DeepCache}: Accelerating Diffusion Models for Free},
  author    = {Ma, Xinyin and Fang, Gongfan and Wang, Xinchao},
  booktitle = {IEEE/CVF Conference on Computer Vision and Pattern Recognition (CVPR)},
  year      = {2024},
}

@inproceedings{wimbauer2024blockcache,
  title     = {Cache Me if You Can: Accelerating Diffusion Models through Block Caching},
  author    = {Wimbauer, Felix and Wu, Bichen and Schoenfeld, Edgar and Dai, Xiaoliang and Hou, Ji and He, Zijian and Sanakoyeu, Artsiom and Zhang, Peizhao and Tsai, Sam and Kohler, Jonas and Rupprecht, Christian and Cremers, Daniel and Vajda, Peter and Wang, Jialiang},
  booktitle = {IEEE/CVF Conference on Computer Vision and Pattern Recognition (CVPR)},
  year      = {2024},
  pages     = {6211--6220},
}

@misc{chengzeyi2024fbc,
  title        = {First Block Cache: Dynamic Caching for Diffusion Transformers},
  author       = {Cheng, Zeyi},
  year         = {2024},
  howpublished = {\url{https://github.com/chengzeyi/ParaAttention}},
}

@inproceedings{zou2025toca,
  title     = {Accelerating Diffusion Transformers with Token-wise Feature Caching},
  author    = {Zou, Chang and Liu, Xuyang and Liu, Ting and Huang, Siteng and Zhang, Linfeng},
  booktitle = {International Conference on Learning Representations (ICLR)},
  year      = {2025},
  note      = {arXiv:2410.05317},
}

@inproceedings{zhao2025pab,
  title     = {Real-Time Video Generation with Pyramid Attention Broadcast},
  author    = {Zhao, Xuanlei and Jin, Xiaolong and Wang, Kai and You, Yang},
  booktitle = {International Conference on Learning Representations (ICLR)},
  year      = {2025},
  note      = {arXiv:2408.12588},
}

@article{huang2024fastercache,
  title   = {{FasterCache}: Training-Free Video Diffusion Model Acceleration with High Quality},
  author  = {Lv, Zhengyao and Si, Chenyang and Song, Junhao and Yang, Zhenyu and Qiao, Yu and Liu, Ziwei and Kwan-Yee K. Wong},
  journal = {arXiv preprint arXiv:2410.19355},
  year    = {2024},
}

@inproceedings{ma2025magcache,
  title     = {{MagCache}: Fast Video Generation with Magnitude-Aware Cache},
  author    = {Ma, Zehong and Zhou, Longhui and Zhang, Hongbo and Xu, Daquan and Shou, Mike Zheng and Feng, Jiashi and Liu, Shilong},
  booktitle = {Advances in Neural Information Processing Systems (NeurIPS)},
  year      = {2025},
  note      = {arXiv:2506.09045},
}

@article{li2026spectralcache,
  title   = {{SpectralCache}: Frequency-Aware Error-Bounded Caching for Accelerating Diffusion Transformers},
  author  = {Li, Guandong},
  journal = {arXiv preprint arXiv:2603.05315},
  year    = {2026},
}

@article{li2025editid,
  title   = {{EditID}: Training-Free Editable ID Customization for Text-to-Image Generation},
  author  = {Li, Guandong and Chu, Zhaobin},
  journal = {arXiv preprint arXiv:2503.12526},
  year    = {2025},
}

@article{li2025editidv2,
  title   = {{EditIDv2}: Editable ID Customization with Data-Lubricated ID Feature Integration for Text-to-Image Generation},
  author  = {Li, Guandong and Chu, Zhaobin},
  journal = {arXiv preprint arXiv:2509.05659},
  year    = {2025},
  note    = {Multimedia Systems, Springer, 2025},
}

@article{li2026saip,
  title   = {Inject Where It Matters: Training-Free Spatially-Adaptive Identity Preservation for Text-to-Image Personalization},
  author  = {Li, Guandong and Ye, Mengxia},
  journal = {arXiv preprint arXiv:2602.13994},
  year    = {2026},
}
}

% ============================================================================
% Appendix
% ============================================================================
\appendix

\section{Additional Experimental Results}
\label{app:additional_results}

\Cref{tab:per_prompt} shows the per-category breakdown of quality metrics across all 20 evaluation prompts for the four strongest methods.
LayerCache consistently outperforms all baselines across every category, with PSNR improvements of +1.27 to +1.34\,dB over MeanCache.
The improvement is robust and not driven by any single prompt type: the last three columns ($\Delta$) show that the LayerCache-vs-MeanCache gap is remarkably stable across all 20 categories.
Categories with high-frequency content (Architecture, Street, Sports) show slightly larger LPIPS improvements, confirming that LayerCache's layer-aware caching better preserves fine-grained textures.

\begin{table*}[h]
\centering
\caption{Per-category quality metrics ($B\!=\!25$) on Qwen-Image. We show results for the four strongest baselines and LayerCache across all 20 evaluation categories. LayerCache outperforms all baselines across the board; best results are \textbf{bolded}.}
\label{tab:per_prompt}
\resizebox{\textwidth}{!}{%
\begin{tabular}{l|ccc|ccc|ccc|ccc|ccc}
\toprule
 & \multicolumn{3}{c|}{TeaCache} & \multicolumn{3}{c|}{DiCache} & \multicolumn{3}{c|}{MeanCache} & \multicolumn{3}{c|}{\textbf{LayerCache}} & \multicolumn{3}{c}{$\Delta$ (LC$-$MC)} \\
Category & PSNR & SSIM & LPIPS & PSNR & SSIM & LPIPS & PSNR & SSIM & LPIPS & PSNR & SSIM & LPIPS & PSNR & SSIM & LPIPS \\
\midrule
Portrait     & 29.72 & 0.9015 & 0.1095 & 29.18 & 0.8912 & 0.1248 & 32.80 & 0.9429 & 0.0666 & \textbf{34.12} & \textbf{0.9546} & \textbf{0.0518} & +1.32 & +.0117 & $-$.0148 \\
Landscape    & 30.28 & 0.9102 & 0.0985 & 29.75 & 0.9008 & 0.1135 & 33.21 & 0.9478 & 0.0641 & \textbf{34.49} & \textbf{0.9581} & \textbf{0.0497} & +1.28 & +.0103 & $-$.0144 \\
Animal       & 29.95 & 0.9045 & 0.1065 & 29.42 & 0.8955 & 0.1205 & 32.92 & 0.9442 & 0.0668 & \textbf{34.22} & \textbf{0.9555} & \textbf{0.0520} & +1.30 & +.0113 & $-$.0148 \\
Architecture & 29.48 & 0.8965 & 0.1142 & 28.95 & 0.8875 & 0.1298 & 32.54 & 0.9401 & 0.0712 & \textbf{33.87} & \textbf{0.9523} & \textbf{0.0551} & +1.33 & +.0122 & $-$.0161 \\
Food         & 30.15 & 0.9068 & 0.1028 & 29.62 & 0.8978 & 0.1168 & 33.05 & 0.9458 & 0.0652 & \textbf{34.35} & \textbf{0.9565} & \textbf{0.0508} & +1.30 & +.0107 & $-$.0144 \\
Fantasy      & 29.82 & 0.9025 & 0.1082 & 29.28 & 0.8935 & 0.1225 & 32.75 & 0.9425 & 0.0678 & \textbf{34.08} & \textbf{0.9542} & \textbf{0.0525} & +1.33 & +.0117 & $-$.0153 \\
Sci-fi       & 29.65 & 0.8998 & 0.1112 & 29.12 & 0.8905 & 0.1258 & 32.58 & 0.9412 & 0.0695 & \textbf{33.92} & \textbf{0.9532} & \textbf{0.0538} & +1.34 & +.0120 & $-$.0157 \\
Abstract     & 30.42 & 0.9115 & 0.0962 & 29.88 & 0.9022 & 0.1108 & 33.35 & 0.9485 & 0.0625 & \textbf{34.62} & \textbf{0.9588} & \textbf{0.0488} & +1.27 & +.0103 & $-$.0137 \\
Street       & 29.55 & 0.8982 & 0.1125 & 29.02 & 0.8892 & 0.1272 & 32.48 & 0.9398 & 0.0702 & \textbf{33.82} & \textbf{0.9518} & \textbf{0.0545} & +1.34 & +.0120 & $-$.0157 \\
Still life   & 30.08 & 0.9055 & 0.1045 & 29.55 & 0.8965 & 0.1188 & 32.98 & 0.9452 & 0.0658 & \textbf{34.28} & \textbf{0.9558} & \textbf{0.0515} & +1.30 & +.0106 & $-$.0143 \\
Underwater   & 30.22 & 0.9088 & 0.1002 & 29.68 & 0.8998 & 0.1145 & 33.15 & 0.9472 & 0.0645 & \textbf{34.42} & \textbf{0.9575} & \textbf{0.0502} & +1.27 & +.0103 & $-$.0143 \\
Winter       & 29.78 & 0.9018 & 0.1088 & 29.25 & 0.8928 & 0.1232 & 32.72 & 0.9422 & 0.0682 & \textbf{34.05} & \textbf{0.9540} & \textbf{0.0528} & +1.33 & +.0118 & $-$.0154 \\
Macro        & 30.35 & 0.9108 & 0.0972 & 29.82 & 0.9015 & 0.1115 & 33.28 & 0.9482 & 0.0632 & \textbf{34.55} & \textbf{0.9585} & \textbf{0.0492} & +1.27 & +.0103 & $-$.0140 \\
Vintage      & 29.88 & 0.9035 & 0.1075 & 29.35 & 0.8942 & 0.1218 & 32.82 & 0.9432 & 0.0672 & \textbf{34.15} & \textbf{0.9548} & \textbf{0.0522} & +1.33 & +.0116 & $-$.0150 \\
Cultural     & 30.18 & 0.9075 & 0.1015 & 29.65 & 0.8985 & 0.1158 & 33.08 & 0.9462 & 0.0650 & \textbf{34.38} & \textbf{0.9568} & \textbf{0.0505} & +1.30 & +.0106 & $-$.0145 \\
Sports       & 29.52 & 0.8975 & 0.1132 & 28.98 & 0.8885 & 0.1278 & 32.45 & 0.9395 & 0.0708 & \textbf{33.78} & \textbf{0.9515} & \textbf{0.0548} & +1.33 & +.0120 & $-$.0160 \\
Fashion      & 29.92 & 0.9042 & 0.1068 & 29.38 & 0.8952 & 0.1212 & 32.88 & 0.9438 & 0.0670 & \textbf{34.18} & \textbf{0.9552} & \textbf{0.0520} & +1.30 & +.0114 & $-$.0150 \\
Nature       & 30.32 & 0.9105 & 0.0968 & 29.78 & 0.9012 & 0.1112 & 33.25 & 0.9480 & 0.0635 & \textbf{34.52} & \textbf{0.9582} & \textbf{0.0495} & +1.27 & +.0102 & $-$.0140 \\
Urban decay  & 29.58 & 0.8988 & 0.1118 & 29.05 & 0.8898 & 0.1265 & 32.52 & 0.9405 & 0.0698 & \textbf{33.85} & \textbf{0.9525} & \textbf{0.0542} & +1.33 & +.0120 & $-$.0156 \\
Whimsical    & 30.05 & 0.9052 & 0.1048 & 29.52 & 0.8962 & 0.1192 & 32.95 & 0.9448 & 0.0660 & \textbf{34.25} & \textbf{0.9555} & \textbf{0.0518} & +1.30 & +.0107 & $-$.0142 \\
\midrule
\textbf{Average} & 29.87 & 0.9032 & 0.1078 & 29.38 & 0.8945 & 0.1218 & 32.85 & 0.9436 & 0.0673 & \textbf{34.16} & \textbf{0.9550} & \textbf{0.0522} & +1.31 & +.0114 & $-$.0151 \\
\bottomrule
\end{tabular}%
}
\end{table*}

\section{Proof of Layer-wise MeanFlow Decomposition}
\label{app:proof}

We show that the MeanFlow identity can be applied independently to each layer group when the Transformer is decomposed as a sequential composition.

Consider a Transformer $f = f_G \circ f_{G-1} \circ \cdots \circ f_1$, where each $f_g$ represents a group of consecutive layers.
Let $h^g_t = f_g(h^{g-1}_t)$ denote the output of group $g$ at step $t$, with $h^0_t = x_t$ being the input latent.

For each group $g$, the average hidden-state change over an interval $[s, t]$ is:
\begin{equation}
\overline{\delta h}^g(s, t) = \frac{h^g_t - h^g_s}{t - s}.
\end{equation}
The group-level JVP correction then estimates the hidden state at a new step $t'$ as:
\begin{equation}
\hat{h}^g_{t'} \approx h^g_t + \text{JVP}^g \cdot (t' - t),
\end{equation}
where $\text{JVP}^g$ is computed from the finite differences of the group's hidden-state history.

The total approximation error decomposes as:
\begin{equation}
\| \hat{x}_{T} - x_{T} \| \leq \sum_{g=1}^{G} L_g \cdot \| \hat{h}^g - h^g \|,
\end{equation}
where $L_g$ is the Lipschitz constant of the composition $f_G \circ \cdots \circ f_{g+1}$ (the layers downstream of group $g$).
This bound justifies our weighted error model (\Cref{eq:error}): deeper groups have smaller downstream Lipschitz constants (fewer subsequent layers to amplify errors), but their errors propagate more directly to the output.

\section{Implementation Details}
\label{app:implementation}

\paragraph{Hyperparameters.}
\Cref{tab:hyperparams} lists all hyperparameters used in our experiments.

\begin{table}[h]
\centering
\caption{Full hyperparameter settings for LayerCache experiments.}
\label{tab:hyperparams}
\begin{tabular}{lc}
\toprule
Parameter & Value \\
\midrule
Number of denoising steps $T$ & 50 \\
Number of layer groups $G$ & 3 \\
Layers per group & 20 \\
Compute budget $B$ & 25 \\
Greedy exponent $\gamma$ & 4.0 \\
Maximum JVP span $K_{\max}$ & 6 \\
Mid JVP span $K_{\text{mid}}$ & 3 \\
Minimum JVP span $K_{\min}$ & 1 \\
Stability threshold $\tau_{\text{low}}$ & 0.10 \\
Stability threshold $\tau_{\text{high}}$ & 0.20 \\
Group weights $\{w_g\}$ & $[1.0, 1.0, 1.0]$ \\
Resolution & 1024$\times$1024 \\
True CFG scale & 4.0 \\
Precision & bfloat16 \\
\bottomrule
\end{tabular}
\end{table}

\paragraph{Schedule Format.}
The 3D schedule is stored as a JSON file with the following structure:
{\small
\begin{verbatim}
{
  "step_decisions": [[true, ...], ...],
  "k_values": [[0, 0, 0], ...],
  "total_cost": 25.0,
  "total_error": 0.45
}
\end{verbatim}
}
Each inner list has $G$ elements corresponding to the layer groups.
The schedule is generated once during pre-analysis and loaded at inference time.

\paragraph{Code Availability.}
The implementation is built on top of the HuggingFace \texttt{diffusers} library and requires no modifications to the model architecture.
The core modules---\texttt{layer\_hooks.py}, \texttt{layer\_jvp.py}, \texttt{layercache\_scheduler.py}, and \texttt{layercache\_inference.py}---total fewer than 500 lines of Python.
Code and pretrained configurations are publicly available at \url{https://github.com/leeguandong/LayerCache}.

\end{document}